\documentclass[10pt,twocolumn,letterpaper]{article}

\usepackage{wacv}
\usepackage{times}
\usepackage{epsfig}
\usepackage{graphicx}
\usepackage{amsmath}
\usepackage{amssymb}
\usepackage{placeins}
\usepackage{tabularx}
\usepackage{booktabs}

\newcommand{\fig}[1]{Figure~\ref{fig:#1}}
\newcommand{\sect}[1]{Section~\ref{sect:#1}}
\newcommand{\tab}[1]{Table~\ref{tab:#1}}

\newcommand{\eq}[1]{(\ref{eq:#1})}

\def\wacvPaperID{1134} 

\wacvfinalcopy 
\ifwacvfinal
\def\assignedStartPage{1} 
\fi

\ifwacvfinal
\usepackage[breaklinks=true,bookmarks=false]{hyperref}
\else
\usepackage[pagebackref=true,breaklinks=true,colorlinks,bookmarks=false]{hyperref}
\fi

\ifwacvfinal
\else
\fi

\begin{document}

\title{Real-time RGBD-based Extended Body Pose Estimation}

\author{
Renat Bashirov $^1$\thanks{equal contribution} \and
Anastasia Ianina $^1$\footnotemark[1] \and
Karim Iskakov $^{1,2}$ \and
Yevgeniy Kononenko $^1$ \and
Valeriya Strizhkova $^2$ \and
Victor Lempitsky $^{1,2}$ \and
Alexander Vakhitov $^1$ \and
\\ $^1$ Samsung AI Center – Moscow, Russia \and
\\ $^2$ Skolkovo Institute of Science and Technology, Russia
}
\maketitle

\begin{abstract}
We present a system for real-time RGBD-based estimation of 3D human pose. We use parametric 3D deformable human mesh model (SMPL-X) as a representation and focus on the real-time estimation of parameters for the body pose, hands pose and facial expression from Kinect Azure RGB-D camera. We train estimators of body pose and facial expression parameters. Both estimators use previously published landmark extractors as input and custom annotated datasets for supervision, while hand pose is estimated directly by a previously published method. We combine the predictions of those estimators into a temporally-smooth human pose. We train the facial expression extractor on a large talking face dataset, which we annotate with facial expression parameters. For the body pose we collect and annotate a dataset of 56 people captured from a rig of 5 Kinect Azure RGB-D cameras and use it together with a large motion capture AMASS dataset. Our RGB-D body pose model outperforms the state-of-the-art RGB-only methods and works on the same level of accuracy compared to a slower RGB-D optimization-based solution. The combined system runs at 30 FPS on a server with a single GPU. The code will be available at \href{https://saic-violet.github.io/rgbd-kinect-pose}{\texttt{saic-violet.github.io/rgbd-kinect-pose}}
\end{abstract}

\newcommand{\bj}{{\bf j}} 
\section{Introduction}

A decade ago, realtime human pose estimation using an RGBD sensor became a landmark achievement for computer vision~\cite{KinectAnnouncement,shotton2012efficient}. Over the subsequent decade, the focus in human pose estimation has however shifted onto RGB sensors~\cite{openpose1,chen2020monocular}. Also, while the original Kinect-based approach and many subsequent RGB-based works aimed at skeleton joints estimation, most recent trend is to estimate \textit{extended} pose description that includes face expression and hands pose~\cite{joo2018total,pavlakos2019expressive}.

Here, we argue that despite all the progress in RGB-based pose estimation, the availability of depth can still be of great use for the pose estimation task. We therefore build an RGBD-based system for realtime pose estimation that uses a modern representation for extended pose (SMPL-X~\cite{pavlakos2019expressive}) involving face and hands pose estimation. To build the system, we adopt a simple fusion approach, which uses pretrained realtime estimators for body, face, and hands poses. We then convert the outputs of these estimators into a coherent set of SMPL-X parameters.

To train our system, we collect a dataset of 56 people using a calibrated rig of five Kinect sensors. We then establish ``ground truth'' poses using slow per-frame optimization-based fitting process that accurately matches multi-view observations. We also fit the deformable head mesh to the videos from the large-scale VoxCeleb2 dataset~\cite{chung2018voxceleb2}. The result of this fitting is then used as a ground truth for the learnable components of our system.

To recover the body pose, we train a neural network that converts the stream of depth-based skeleton estimates (as provided by the Kinect API~\cite{BodyTracker}) into a stream of SMPL pose parameters (angles). We use a specific (residual) parameterization, which asks the network to predict corrections to the angles of the Kinect Skeleton.
In the comparison, we observe that tracking accuracy of such depth-based system considerably exceeds the accuracy of the state-of-the-art RGB-based methods~\cite{ExPose:2020,kolotouros2019learning,kocabas2019vibe}, validating the usefulness of the depth channel for human pose estimation. Furthermore, we compare the performance of our feed-forward network with the depth-based baseline that performs per-frame optimization of pose parameters. We observe that the feed-forward network achieves same accuracy and much higher speed.

In addition to the body parameters inferred from depth channel, we estimate the face and the hand parameters from the RGB color stream, since the effective resolution of the depth channel at the camera-to-body distances typical for fullbody tracking is not sufficient for these tasks. We use the outputs of the pretrained MinimalHand system~\cite{zhou2020monocular} to perform SMPLX-compatible (MANO~\cite{romero2017embodied}) hand pose estimation. We also estimate the face keypoints using the MediaPipe face mesh model~\cite{kartynnik2019real}, and train a feedforward network that converts the keypoint coordinates into the SMPL-X compatible (FLAME~\cite{li2017learning}) face parameters. The MANO and FLAME estimates are fused with the body pose to complete the extended pose estimation.

Our resulting system thus uses the depth channel to estimate the body pose, and the color channel to estimate hands and face poses. It runs at 25 frames-per-second on a modern desktop with a single 2080TI GPU card, and provides reliable extended body tracking. We demonstrate that same as ten years back, RGBD-based pose estimation still strikes a favourable balance between the simplicity of the setup (single sensor is involved, no extrinsic calibration is needed), and the accuracy of pose tracking (in particular, higher accuracy is attainable compared to monocular RGB-based estimation).

 \section{Related work}

Classic methods approach human pose estimation as sparse keypoint localization in color~\cite{ramanan2006learning,ferrari2008progressive,felzenszwalb2008discriminatively} or RGB-D~\cite{ye2011accurate,shotton2012efficient} images. In~\cite{taylor2012vitruvian} the authors use dense per-pixel prediction of surface coordinates in RGB-D images. Deep network-based methods are significantly more accurate in such a  direct regression, e.g.~\cite{toshev2014deeppose,newell2016stacked,sun2019deep}. Some methods estimate the body joints in 3D from single~\cite{martinez2017simple} or multiple RGB views~\cite{iskakov2019learnable}. RGBD-based 3D joint prediction using deep learning is proposed in~\cite{Martinez_IROS_2020,moon2018v2v,haque2016towards}, and in this work we build on a recent commercial system~\cite{BodyTracker} of this kind. In~\cite{zhou2016deep}, Zhou~et~al.~use known constant bone lengths and predict joint angles of a body kinematic model, for the first time proposing a deep architecture which estimates the body pose assuming the body shape is known in the form of bone lengths. Subsequent research~\cite{mehta2017vnect,sun2017compositional,sun2018integral,arnab2019exploiting} focuses on incorporating structural knowledge into pose estimation. 

Parametric body models~\cite{loper2015smpl,pavlakos2019expressive,joo2018total,xu2020ghum,Hesse:MICCAI:2018} are  rich and higher level representations of body geometry separated into pose and person-specific shape. In this work we rely on a recent SMPL-X model~\cite{pavlakos2019expressive} which integrates body, face and hand models into a unified framework. The most accurate methods for the estimation of the body model parameters are computationally intensive and offline~\cite{bogo2016keep,pavlakos2019expressive,xiang2019monocular,Weiss:ICCV:11,Bogo:ICCV:2015,PROX:2019}. Among them ~\cite{Weiss:ICCV:11,Bogo:ICCV:2015,PROX:2019} take depth data from Kinect sensor into account to reconstruct parametric body model using non real-time optimization. The initial work~\cite{Weiss:ICCV:11} demonstrated that a single inexpensive commodity sensor could achieve registration accuracy similar to systems based on multiple calibrated cameras and structured light sources. The follow-up work~\cite{Bogo:ICCV:2015} shows laser-scan-quality body reconstruction results for RGB-D sequences of freely moving people using multi-scale parametric models augmented with displacement maps. Most recently~\cite{PROX:2019} adds environmental constraints enforcing the consistency between the fitting results and a 3D scene model. 

Recent research features feedforward deep architectures achieving real-time prediction of the body model parameters~\cite{kanazawa2018end,kolotouros2019learning,ExPose:2020}. In particular, \cite{kanazawa2018end} proposes to estimate shape and pose parameters of the SMPL model from a single RGB image, while recent works, e.g.~SPIN~\cite{kolotouros2019learning}, show significant accuracy improvements in this direction. ExPose~\cite{ExPose:2020} is a real-time capable deep network predicting the whole set of SMPL-X parameters. VIBE~\cite{kocabas2019vibe} is a fast video-based SMPL shape and pose prediction method, while~\cite{xu2020ghum} shows a similar approach for the newer GHUM model.

For the body models, the recent research on body pose forecasting~\cite{aksan2019structured,aksan2020attention} relies on a large motion capture dataset~\cite{mahmood2019amass} with SMPL annotations. However, these approaches deal with prediction of noise-free poses rather than extraction of the high-level pose representation in the form of angles from the lower-level representation, as in this work. Another closely related field is skeletal retargeting, where researchers find ways of transforming the skeletons between different animated characters, trying to keep the pose semantics intact, e.g.~\cite{aberman2019,aberman2020skeleton}.

This work proposes a sequence-based  method to estimate poses of the SMPL-X model, which assumes knowledge of the body shape as opposed to VIBE~\cite{kocabas2019vibe} and~\cite{zanfir2020weakly}, and uses a sequence of RGB-D frames as input.

 \section{Methods}

\begin{figure}
    \centering
    \includegraphics[width=0.5\textwidth]{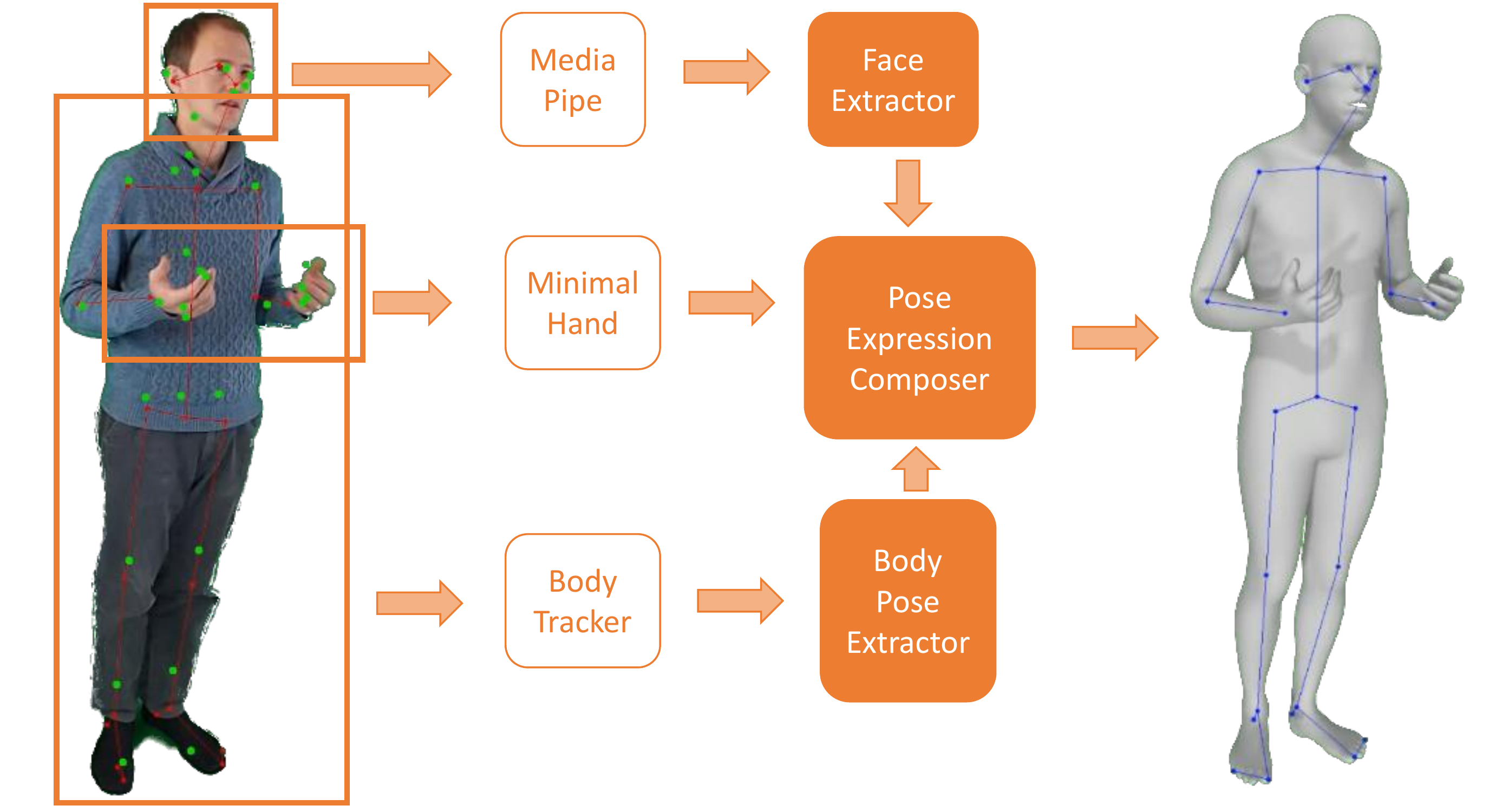}
    \caption{The real-time system predicts the SMPL-X~\cite{pavlakos2019expressive} pose and expression from the RGB-D frame. We rely on the Kinect Body Tracker~\cite{BodyTracker}, the MediaPipe face mesh~\cite{kartynnik2019real} and the hand pose regressor~\cite{zhou2020monocular} (empty boxes). We propose the {\em body pose extractor} to estimate the body joint angles from the 3D landmarks, the {\em face extractor} to estimate the facial expression and jaw pose, and the {\em pose expression composer} to integrate the predictions into a single set of consistent SMPL-X parameters (filled boxes).}
    \label{fig:scheme}
    \vspace{-0.5cm}
\end{figure}

\begin{figure*}
    \centering
    \includegraphics[width=\textwidth]{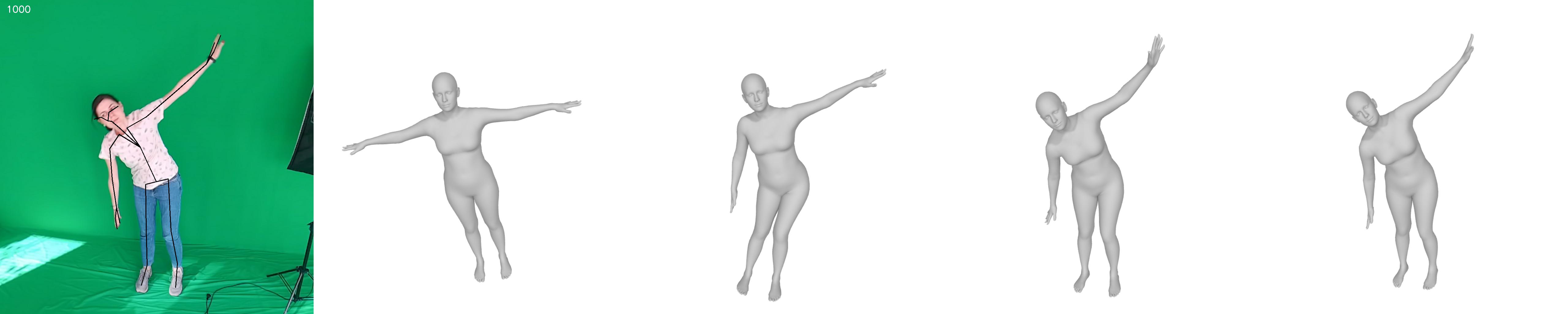}
    \begin{tabular}{ccccc}
    \quad  input image \quad \qquad & 
    \qquad +global rotation \qquad &
    \quad +bending heuristic \quad &
    \quad +refinement by GRU \quad &
    \qquad ground truth \qquad
    \end{tabular}
    \caption{Our body pose extractor has 4 steps. Step 1: align the SMPL mesh with a global rotation. Step 2: apply learning-free bending heuristic to the limbs and the neck. Steps 3 and 4: refine the joint rotations with a GRU network, refine global rotation and translation.}
    \label{fig:steps}
    \vspace{-0.5cm}
\end{figure*}

We use the SMPL-X (Skinned Multi-Person Linear - eXpressive) format for extended pose~\cite{pavlakos2019expressive}, which represents the shape of a  human body using parametric deformable mesh model. The mesh vertex positions $M$ are determined by a set of body shape $\beta$, pose $\Theta$ and facial expression $\psi$ parameters as follows:
\begin{equation}\label{eq:smplx}
    \begin{split}
        M(\beta, \Theta, \psi) &= W(T_p(\beta, \Theta, \psi), J(\beta), \Theta, {\cal W})\\
        T_p(\beta, \Theta, \psi) &= {\bar T} + B_S(\beta; {\cal S}) + B_E(\psi; {\cal E}) + B_P(\Theta; {\cal P})\,.
    \end{split}
\end{equation}
Here, $M(\beta, \Theta, \psi)$ is the posed human mesh, $T_p(\beta, \Theta, \psi)$ encodes the deformed mesh in a default body pose,  $W$ is a linear blend skinning function with vertex-joint assignment weights ${\cal W}$ and joint coordinates $J(\beta)$; the mesh $T_p$ is expressed as a sum of the vector of mean vertex coordinates ${\bar T}$ summed with the blend-shape function $B_S$, the displacements corresponding to the facial expression $B_E$ and the pose $B_P$, where ${\cal S}$, ${\cal E}$, ${\cal P}$ denote the mesh deformation bases due to shape, expression and pose, see~\cite{pavlakos2019expressive} for the full definition of the model. The pose can be decomposed as $\Theta = [\theta^T, \, \xi^T, \, \zeta^T]^T$ into the parts corresponding to the hands $\xi$, the face $\zeta$ and the remaining body joints $\theta$. The SMPL-X model is an ``amalgamation'' of the SMPL body model~\cite{loper2015smpl}, the FLAME face/head model~\cite{li2017learning}, and the MANO hand model~\cite{romero2017embodied} that had been proposed earlier.

\subsection{System Overview}

The proposed system regresses the  sequence of the pose-expression pairs $(\Theta_i, \psi_i)$ from a sequence of RGB-D frames, see Fig.~\ref{fig:scheme}. As a first step of processing the frame $i$ it extracts the vector of 3D body landmark predictions $\bj_i$ using the Kinect Body Tracker API~\cite{BodyTracker}. For simplicity, we assume that they are defined in the RGB camera coordinates, in reality we use the intrinsics provided with the device for the coordinate system alignment. We predict the body pose sequence $\{\theta_{i}\}_{i=1}^N$, where $N$ is the total number of frames, from a sequence of 3D landmarks $\{\{x_{i,k}\}_{k=1}^{m}\}_{i=1}^N$ using our {\em body pose extractor} module, where $x_{i,k}$ is the $k$-th 3D landmark for the $i$-th frame. We then crop the face and process it with MediaPipe face mesh predictor~\cite{kartynnik2019real}, and later extract the facial expression $\psi_i$ and the pose of the jaw $\zeta_{ i}$ using our {\em facial extractor} module. We crop the hands and predict the hand pose $\xi_{i}$ in the SMPL-X (MANO) format using the MinimalHand~\cite{zhou2020monocular} method. The components of the model are then combined together, and temporal smoothing is applied: Slerp Quaternion interpolation is used for body, jaw and hands rotations, and exponential temporal filter is used for face expression.

\begin{table*}[htbp]
    \centering
    \begin{tabularx}{\textwidth}{||c|X|X|X|X|X|X|X|X|X|X|X|X||}
         \hline
         & \multicolumn{4}{|c|}{simple} & \multicolumn{4}{c|}{complex} 
         & \multicolumn{4}{c||}{avg} \\
         \hline
          & AUC$\uparrow$ & Euler$\downarrow$ & joint-ang$\downarrow$ & Pos$\downarrow$ & AUC$\uparrow$ & Euler$\downarrow$ & joint-ang$\downarrow$ & Pos$\downarrow$
          & AUC$\uparrow$ & Euler$\downarrow$ & joint-ang$\downarrow$ & Pos$\downarrow$\\
         \hline
         \hline
         \multicolumn{13}{||c||}{single-frame} \\
         \hline
         SMPLify-X~\cite{pavlakos2019expressive} 
         & 0.735 & {\bf 2.680} & 0.324 & 0.079 & 0.651 & 2.224 & 0.474 & 0.136 
         & 0.693 & 2.452 & 0.399 & 0.108 \\
         \hline
         SPIN~\cite{kolotouros2019learning}
         & 0.802 & 3.018 & 0.298 & 0.058 & 0.756 & 1.920 & 0.342 & 0.074
         & 0.779 & 2.469 & 0.32 & 0.066 \\
         \hline
         ExPose-X~\cite{ExPose:2020}
         & 0.781 & 3.345 & 0.305 & 0.064 & 0.746 & 2.6 & 0.369 & 0.078
         & 0.764 & 2.972 & 0.337 & 0.071 \\
         \hline
         Ours MLP 
         & {\bf 0.853} & 2.878 & {\bf 0.240} & {\bf 0.040} 
         & {\bf 0.777} & {\bf 1.552} & {\bf 0.296} & {\bf 0.068}
         & {\bf 0.815} & {\bf 2.215} & {\bf 0.268} & {\bf 0.054}\\
         \hline
         \multicolumn{13}{||c||}{multi-frame} \\
         \hline
         VIBE~\cite{kocabas2019vibe} 
         & 0.823 & {\bf 2.744} & 0.271 & 0.052 
         & 0.753 & 1.763 & 0.323 & 0.076
         & 0.788 & {\bf 2.254} & 0.297 & 0.064 \\
         \hline
         Ours RNN 
         & 0.870 & 2.988 & 0.233 & 0.038 
         & 0.791 & 1.846 & 0.287 & 0.063
         & 0.83 & 2.417 & 0.26 & 0.05 \\
         \hline
         Ours RNN-SPL 
         & {\bf 0.877} & 3.050 & {\bf 0.226} & {\bf 0.036} 
         & {\bf 0.810} & {\bf 1.735} & {\bf 0.260} & {\bf 0.057}
         & {\bf 0.844} & 2.392 & {\bf 0.243} & {\bf 0.046}\\
         \hline
    \end{tabularx}
    \caption{Comparison with single- and multi-frame RGB baselines on AzurePose (simple/complex). For all metrics except AUC, lower is better. Our RGBD-based methods are more accurate in most metrics (except Euler angles measured for local rotations in simple sequences). }
    \label{tab:azure_test_rgb}
    \vspace{-0.5cm}
\end{table*}

\subsection{Body Pose Extractor}
The aim of the body pose extractor is to predict the body pose $\theta$ given a known SMPL-X body shape vector $\beta$ and the 3D body landmarks $\{x_k\}_{k=1}^{m}$ extracted by the Body Tracker~\cite{BodyTracker} observed at a certain moment of time. The pose extractor should work for an arbitrary shape of a person, while in the public domain to the best of our knowledge there is no RGB-D dataset with sufficient variability of human shapes. We achieve generalization to an arbitrary human shape by two means. Firstly, using the landmark-anchored vertices precomputed for the Body Tracker landmarks, we are able to leverage a large and diverse motion capture dataset~\cite{aksan2019structured} for learning our models. Secondly, we propose a specific residual rotation-based architecture designed to abstract from a particular human shape.
Apart from the temporal connections inside the architecture, the whole pipeline estimates the pose independently for each frame. We therefore omit the temporal indices.

We assume that the Azure Kinect 3D landmark locations $\{x_k\}_{k=1}^m$ are given in the coordinate frame of the RGB camera. For each $x_k$ we create an additional vertex in the SMPL-X mesh: a total of 32 vertices are added to 10475 SMPL-X vertices. For the SMPLX-X mesh aligned with the Azure Kinect skeleton, these additional vertices should coincide with the corresponding $\{x_k\}_{k=1}^m$. Each additional vertex is created to minimize the distance to the $x_k$ over a training dataset as explained at the end of the ~\sect{annotation}. We call these additional vertices \textit{landmark-anchored} vertices. They were created during AzurePose dataset annotation. As the body shape and the pose varies, landmark-anchored vertices follow the SMPL-X skinning equations~\eq{smplx}.

For a landmark $x_k$, we denote the position of the corresponding landmark-anchored vertex in the SMPL-X pose $\theta$ as ${\hat x}_k(\theta)$. Our goal is essentially to find the body pose $\theta$ that aligns landmark-anchored vertices $\{{\hat x}_k(\theta)\}_{k=1}^m$ on the SMPL-X body mesh with the corresponding observed landmarks $\{x_k\}_{k=1}^m$. The body pose extractor takes the observed landmarks and outputs the body pose, while SMPL-X inference takes the body pose and outputs the landmark-anchored vertices, this way the body pose extractor "inverts" the inference of SMPL-X. For body pose extractor we use feed-forward computations without inverse kinematics-like optimizations.

To achieve this, we propose a four-step approach (\fig{steps}).
Firstly, we find the global (rigid) rotation of the body mesh in the default SMPL pose. We define the vertical direction $v$ between the 'chest' and 'pelvis' landmarks and the horizontal direction $w$ between the left and right shoulder landmarks, while the corresponding directions in the mesh with the default pose are ${\hat v}$ and ${\hat w}$, and all the direction vectors have a unit norm. We find the rotation matrix $\mathtt{R}$, such that $v = \mathtt{R} {\hat v}$, $v \times w = \frac{1}{\| {\hat v} \times {\hat w} \|}\mathtt{R} ({\hat v} \times {\hat w})$ which aligns $v$ and ${\hat v}$, and the plane defined by $v$ and $w$ with the one defined by ${\hat v}, {\hat w}$. Such a rotation is unique given that $v$ and $w$ are not collinear (which they are not by construction). It aligns the coordinate system of the root joint with the 3D landmark skeleton.

In the second step, we apply a learning-free bending heuristic to obtain an initial pose estimate $\theta^0$. As Azure Kinect skeleton topology roughly matches SMPL-X skeleton topology, and some of the landmark locations of Azure Kinect skeleton are close to the joints of SMPL-X skeleton, we therefore can set a subset of rotations of SMPL-X joints to match the rotations of Azure Kinect skeleton. To do that we compute two types of bone vectors for Azure Kinect skeleton: the observed bone vectors $b_k = x_k - x_{p(k)}$, where a function $p(k)$ returns the index of the parent of a landmark $k$ in the kinematic tree, and the estimated bone vectors defined by the current estimate of the pose $\theta$: ${\hat b}_k(\theta) = {\hat x}_k(\theta) - {\hat x}_{p(k)}(\theta)$. 
For each landmark-anchored vertex ${\hat x}_k$ we define a \textit{dominating} SMPL-X joint $s(k)$ as a joint with the maximal weight for this particular vertex in the linear SMPL deformable model. 
For two vectors $a$ and $b$, denote as $\mathtt{R}(a, b)$ the matrix of minimal rotation between $a$ and $b$, which we define as a rotation with the axis $\frac{1}{\|a \times b \|} a \times b$ which aligns the vectors $a$ and $b$, such that $\frac{1}{\| a \|} \mathtt{R} a = \frac{1}{\| b\| } b$. We then process a subset of bones ${\cal D}$ corresponding to the limbs and the neck, each bone from $b_k \in {\cal D}$ having a unique $s(p(k))$, and
for each bone $b_k \in {\cal D}$, we set the rotation matrix of $s(p(k))$ to $\mathtt{R}\left({\hat b}_k(\theta),b_k\right)$. We denote the resulting pose $\theta^0$. 
The positions of the vertices in SMPL mesh are affected by multiple joints, and not all of the bones are in ${\cal D}$. 
Thus the pose $\theta^0$ does not result in collinearity of all the bone vectors $\{b_k\}$ and $\{{\hat b}_k(\theta^0)\}$, and the goal of the next step is to achieve a more precise alignment.

In the third step, we use machine learning to get a more accurate alignment. We compute the \textit{residual} minimal rotations $\mathtt{R}_k = \mathtt{R}({\hat b}_k(\theta^0), b_k)$ aligning every estimated bone direction ${\hat b}_k(\theta^0)$ with the observed one $b_k$, in the coordinate frame of the joint $s(p(k))$. Concatenating $l=m-1$ rotation matrices $\mathtt{R}_{k}$, we obtain a $9\times{}l$-dimensional rotation residual vector. This vector serves as an input to a deep network, together with concatenated rotation matrices generated by the initialization heuristic from $\theta^0$, in \tab{different_input_data} we compare this approach to other possible inputs. The goal of this network is thus to refine the estimates produced by the initialization bending heuristic.  

We use Gated Recurrent Unit (GRU) architecture~\cite{cho2014learning} for this refinement network. The network thus predicts incremental rotations $\mathtt{\tilde R}_k$ for each joint of the SMPL-X model, and the resulting $\theta^1$ is obtained as a composition of these rotations with  $\theta^0$:
$\mathtt{R}_k = \mathtt{R}_k^0 \mathtt{\tilde R}_k^1$,
where $\mathtt{R}_k^1$ is the predicted rotation from the $k$-th joint to its parent, $\mathtt{R}_k^0$ is the corresponding rotation decoded from the initialization $\theta^0$. We encode the obtained predictions into the vector $\theta^1$. 

\begin{table*}[htbp]
    \centering
    \begin{tabularx}{\textwidth}{||c|X|X|X|X|X|X|X|X|X|X|X|X||}
         \hline
         & \multicolumn{4}{|c|}{simple} & \multicolumn{4}{c|}{complex} 
         & \multicolumn{4}{c||}{avg} \\
         \hline
          & AUC$\uparrow$ & Euler$\downarrow$ & joint-ang$\downarrow$ & Pos$\downarrow$ & AUC$\uparrow$ & Euler$\downarrow$ & joint-ang$\downarrow$ & Pos$\downarrow$
          & AUC$\uparrow$ & Euler$\downarrow$ & joint-ang$\downarrow$ & Pos$\downarrow$\\
         \hline
         \hline
         SMPLify-RGBD-Online &
         0.897 &2.9 & 0.227 & 0.03 & 0.841 & 1.698 & 0.277 & 0.047 &  0.869 & 2.299 & 0.252 & 0.039
          \\
          \hline
          SMPLify-RGBD & {\bf 0.905} & {\bf 1.821} & {\bf  0.141} & {\bf 0.027} & {\bf  0.887} & {\bf 1.281} & {\bf 0.163} & {\bf  0.033} & {\bf 0.896} & {\bf 1.551} & {\bf 0.152} & {\bf 0.03}  \\
          \hline
         Ours-RNN & 0.899 & 2.846 & 0.177 & 0.029 & 0.829 & 1.514 & 0.237 & 0.051 & 0.864 & 2.18 & 0.207 & 0.04 \\
         \hline
         Ours-RNN-SPL &  0.896 & 2.819 & 0.189 & 0.029 & 0.837 &  1.361 & 0.233 & 0.048 & 0.867 & 2.09 & 0.211 & 0.039 \\
         \hline
    \end{tabularx}
    \caption{Comparison with multi-frame RGB-D baselines on AzurePose (simple/complex), the slow offline optimization-based method is the most accurate. The proposed Ours-RNN-SPL method performs on a similar level of accuracy as the online optimization-based method, but is 2.5 times faster. The feedforward network inside our approach thus serves as an efficient approximation of the optimization process.}
    \label{tab:azure_test_rgbd}
    \vspace{-0.5cm}
\end{table*}

We have tried two variants of the architecture. The first variant utilizes GRU architecture with two layers and hidden size equal to 1000. In the second variant we apply a structured prediction layer (SPL)~\cite{aksan2019structured} with hidden size $64$ to GRU outputs. We use dense SPL which means that while making prediction for a joint we take into account all its ancestors in the kinematic tree instead of using only one parent joint. Also in this architecture we use dropout with rate $0.5$.
The networks are trained by minimizing the $l^1$ loss on concatenated rotation matrices using Adam optimizer~\cite{kingma2014adam} with learning rate $0.0001$. We decided to use the recurrent architecture that uses temporal context from previous frames to predict the corrections. This allows to improve temporal stability (slightly). In the single-frame experiments below, we replace the GRU network with a simple multi-layer perceptron (MLP) with 5 layers (each layer consists of 512 neurons).

In the fourth and final step, we modify the global location and orientation of the resulting mesh through Procrustes analysis between the two sets of 3D points  $\{x_k\}_{k=1}^{m}$ and $\{{\hat x}_k(\theta_i)\}_{k=1}^{m}$. This step further improves the alignment accuracy. 

One may question whether our multi-stage alignment procedure can be replaced with a simpler one, or if the shape body parameters $\beta$ can help at the fine-tuning prediction stage. In the experiments, we therefore provide an ablation study that compares our full procedure described in this sections with baselines.

\subsection{Face and hands}\label{sect:face_and_hands}
For each input frame, we crop the RGB image regions corresponding to the face and the hands. For each hand we select a Body Tracker landmark corresponding to a hand, define a cube with a center in a selected landmark and a side of $0.3m$, project its vertices to image plane, and  determine bounding box from projected points. To extract face region we project Body Tracker 3D face landmarks to the image, calculate minimal square bounding box and expand it by a factor of two.

On the face region, we run the MediaPipe face mesh predictor~\cite{kartynnik2019real}, which outputs $468$ $2.5D$ landmarks and rotated bounding box in real time. Absolute $X$, $Y$-coordinates of  $2.5D$ landmarks are normalized on width and height of the rotated bounding box correspondingly, while relative $Z$-coordinate is scaled with a constant value of $1/256$. Normalized landmarks are used to predict jaw pose and facial expression of the SMPL-X head (FLAME) model. Landmarks are first passed through a seven-layer MLP with \textit{Linear-BatchNorm-ReLU}~\cite{ioffe2015batch, nair2010rectified} blocks to extract 32-dimensional feature vector. Then this feature vector is fed into two linear layers, which predict jaw pose and facial expression separately. The model is trained on a large dataset of video sequences depicting talking humans, which we annotated with SMPL-X parameters as described in the next section. During training we use two losses: Mean Squared Error (MSE) of facial expressions and Mean Per-Joint Position Error (MPJPE) of SMPL-X 3D face keypoints. Additionally, we increase the weight of the mouth 3D keypoints in the MPJPE loss to make the predicted mouth more responsive (see section~\ref{sect:experiments} for ablation study). All face predictors are trained with ADAM optimizer~\cite{kingma2014adam} with a learning rate $0.0005$.

On the hands regions, we run a method~\cite{zhou2020monocular}, which outputs SMPL-X compatible set of pose parameters. 

\subsection{AzurePose dataset and SMPL-X Annotation}\label{sect:annotation}

For training our models, we collect a dataset of 56 subjects recorded  by five synchronous Kinect Azure devices. The maximal horizontal angular parallax between different devices is approximately 90 degreees, the distance to the person from the device is 2-3 meters. Subjects perform a fixed set of motions, with each recording being 5-6 minutes long. As a test set, we collect additionally two recordings of two subjects, with fixed sets of motions, so that the first recording contains basic motions ('simple'), while the other has more occlusions and extreme rotations for several joint ('complex'). The test recordings were taken in a separate session with modified camera geometry to ensure a realistic gap between the train and the test set.

To obtain ground truth SMPL-X poses, we use a slow optimization-based multiview fitting procedure. We thus optimize the SMPL-X parameters to fit the observations. Essentially, we are extending the SMPLify-X~\cite{pavlakos2019expressive} method to process multiview synchronous RGB-D sequences. We use the OpenPose landmarks~\cite{openpose1, simon2017hand} for body, face and hands, as well as 3D landmarks of the Body Tracker~\cite{BodyTracker}, and optimize the smooth $l^1$ regression cost~\cite{girshick2015fast}. We perform body pose estimation in the VPoser domain~\cite{pavlakos2019expressive}, use the $l^2$ costs for regularizing the estimation of the jaw and eye pose in the angular domain, the hand pose in the MANO~\cite{romero2017embodied} parameter domain, the body shape in the SMPL-X domain, face expression in the FLAME~\cite{li2017learning} domain, the body pose increment between the subsequent frames in the VPoser~\cite{pavlakos2019expressive} domain.

\begin{figure}
    \centering
    \includegraphics[width=0.49\textwidth]{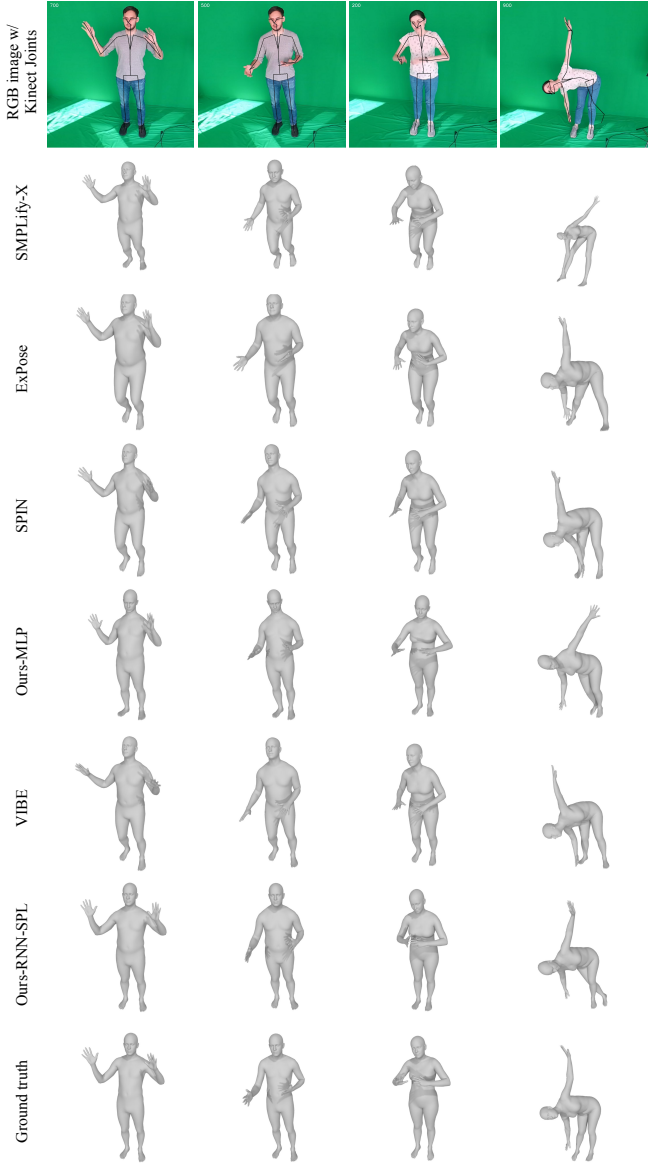}
    \caption{Comparison of the RGB baselines and the proposed methods on AzureTest frames. The use of depth in the proposed methods in most cases leads to more accurate  poses. The rightmost column shows a failure case where our method cannot recover from a mistake of the Kinect BodyTracker.}
    \label{fig:rgb_samples}
    \vspace{-0.5cm}
\end{figure}

\begin{figure}
    \centering
    \includegraphics[width=0.46\textwidth]{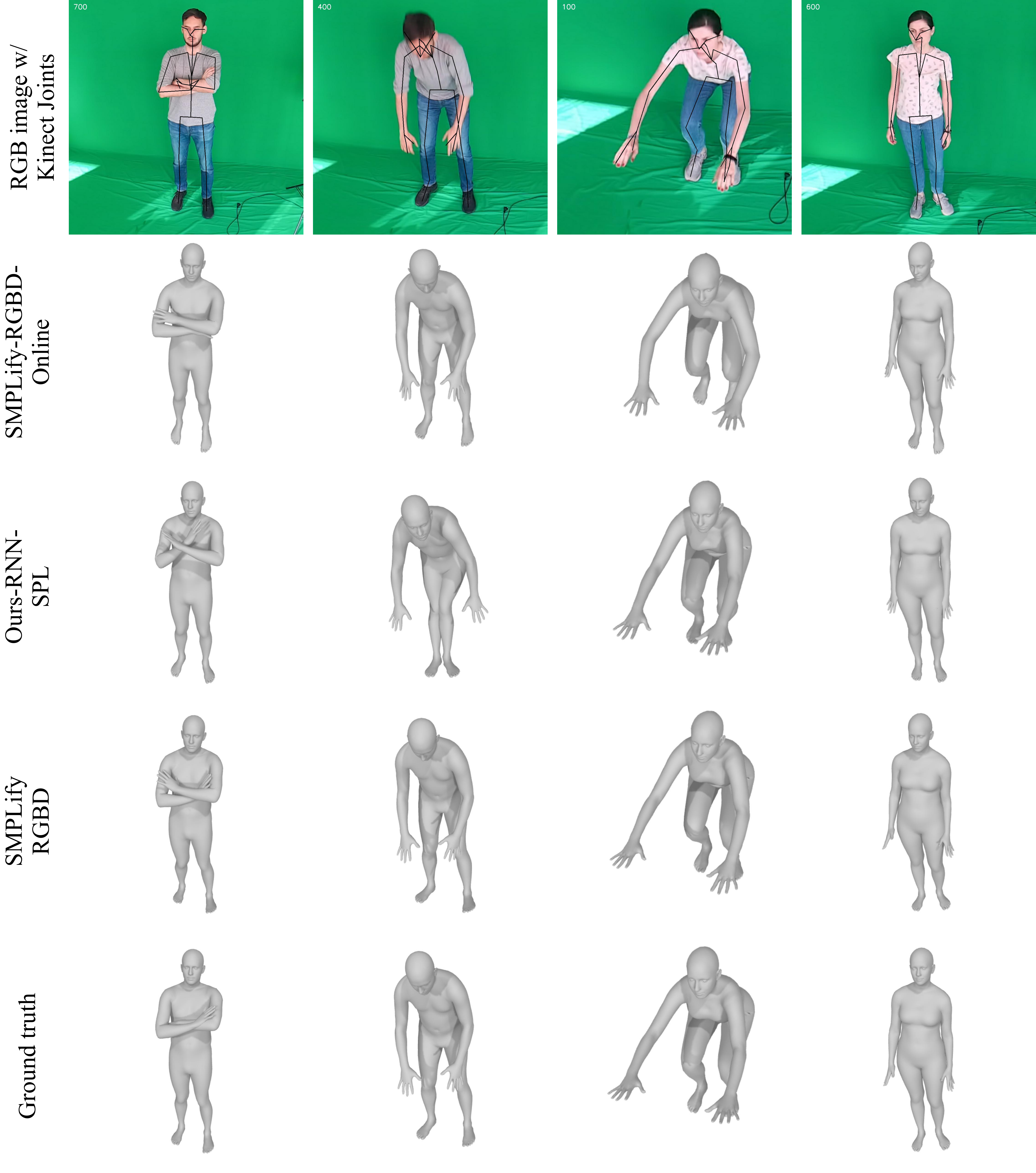}
    \caption{Comparison of the RGB-D baselines and the proposed methods on AzureTest frames. The second column shows an example where our method performs worse than optimization-based approaches. In other examples, the proposed Ours-RNN-SPL method matches the optimization methods closely. 
    }
    \label{fig:rgbd_samples}
    \vspace{-0.5cm}
\end{figure}

We use the explained procedure to obtain SMPL-X annotations on the AzurePose dataset. Since the AzurePose dataset has limited identity and pose variability, we have also built a (semi)-synthetic test set based on AMASS dataset~\cite{mahmood2019amass}, which contains a large variety of tracks in the SMPL format captured with motion capture equipment. We turned the AMASS tracks into synthesized keypoint tracks by taking the coordinates of the landmark-anchored vertices and treating them as Kinect body tracker output. We call the obtained dataset 'AMASS-K'.

To enable feedforward prediction of facial pose and expression, we also require a large dataset of face videos annotated with facial parameters of SMPL-X, which are essentially the parameters of the FLAME model. We use a VoxCeleb2 dataset~\cite{chung2018voxceleb2} for this purpose. First, we select sequences in which the speaker's face has a high resolution (face bounding box more than $512\times 512$ pixels). Then OpenPose~\cite{openpose1} is used to get face landmarks, and the subset of sequences is filtered again according to landmark's confidences. Next, we use an offline optimization-based sequence fitting procedure. Specifically, we optimize the shape (shared across sequence), expression, and jaw parameters of the SMPL-X model to fit the observations. We use the smooth $l^1$ regression cost for optimization and also add $l^2$ cost for smoothness regularization.

For our tasks, we require the landmark-anchored vertices, for which the average distance between them and corresponding Body Tracker landmarks is minimal. We find the optimal landmark-anchored vertices consistently with linear blend skinning (LBS) transformations~(\ref{eq:smplx}). Firstly, we find a closest SMPL-X vertex $i$ to a landmark over the SMPL-X annotated dataset, denoting its LBS weights as ${\cal W}_i$. Then, we optimize for the coordinates of a new virtual vertex $v_m$ in a default body pose, fixing the LBS weights of this vertex equal to ${\cal W}_i$. Finally, we add a vertex with weights ${\cal W}_m = {\cal W}_i$ and coordinates $v_m$ to the model.

\begin{table*}[htbp]
    \centering
    \begin{tabularx}{\textwidth}{||c|X|X|X|X|X|X|X|X|X|X|X|X|X||}
         \hline
         & 
         & \multicolumn{4}{c|}{Input type}
         & \multicolumn{4}{c|}{AMASS-K}
         & \multicolumn{4}{c||}{AzurePose Test (simple/complex)} 
         \\
         \hline
         
         Method 
         & Res
         & KP & $\beta$ & Init & Twists 
         & AUC$\uparrow$ & Euler$\downarrow$ & joint-ang$\downarrow$ & Pos$\downarrow$ 
         & AUC$\uparrow$ & Euler$\downarrow$ & joint-ang$\downarrow$ & Pos$\downarrow$ \\
         \hline
         \hline
         0 &- &+ &- &- &- & 0.885 & 1.629 & 0.211 & 0.033 & 0.849 & 2.187 & 0.231 & 0.046 \\
         \hline
         1 &- &+ &+ &- &- & 0.859 & 1.761 & 0.238 & 0.041 & 0.852 & 2.204 & 0.237 & 0.045 \\
         \hline
         2 &- &+ &- &+ &- & 0.9 & 1.492 & 0.181 & 0.028 & 0.863 & 2.16 & 0.213 & 0.04 \\
         \hline
         3 &+ &- &- &- &+ & 0.935 & 3.001 & 0.139 & 0.018 & 0.761 & 3.897 & 0.443 & 0.075\\
         \hline

         4 &+ &+ &+ &+ &- & 0.897 & 1.543 & 0.184 & 0.029 & 0.864 & {\bf 2.118} & 0.215 & 0.04 \\
         \hline
         5 &+ &- &- &+ &+ & {\bf 0.937} & {\bf 1.264} & {\bf 0.109} & {\bf 0.017} & {\bf 0.864} & 2.18 & {\bf 0.207} & {\bf 0.04} \\
         \hline
    \end{tabularx}
    \caption{Comparison of our models, with different types of inputs and outputs, see text. A model taking the initial pose $\theta_i^0$ and the minimal bone-aligning rotations, and producing the rotation increments is the best on the shape-diverse AMASS-K dataset.}
    \label{tab:different_input_data}
    \vspace{-0.5cm}
\end{table*}
\section{Experiments}{\label{sect:experiments}}
In this section, we evaluate the accuracies of our body pose and face extractors, and report timings of the obtained system. Additional qualitative results and comparisons are available in the \textbf{Supplementary video}. 

{\bf Datasets.} Firstly, we use the test set of the AzurePose dataset, collected by ourselves, see \sect{annotation}.
As this dataset has low shape variability (just two people), we also report test results on the hold-out part of the AMASS-K dataset (140 sequences with 1120 frames).

To measure the quality of the face predictor we filtered and annotated the test split of VoxCeleb2 dataset~\cite{chung2018voxceleb2}, which consists of subjects, who don't appear in the train set. For evaluation, we randomly selected $150$ annotated test videos with $73$ unique subjects. Our face evaluation set covers a wide variety of human face appearance, camera viewpoints, and motions.
 
{\bf Metrics.} We use the set of metrics proposed by recent SMPL pose prediction work~\cite{aksan2019structured}. We obtain the {\em Euler angles} in the $z,y,x$ order, choose the solution with the least amount of rotation, and compute a mean over the $l^2$ norm of the vectors of concatenated angles for all body joints. Next, we compute the {\em joint angle difference} as the mean $l^2$ norm of the rotation logarithm for each body joint. Following~\cite{aksan2019structured}, we compute the Euler angle error on local rotations of joints with respect to their kinematic parents, and the joint angle difference on global rotations between the joint and the body root coordinate systems. We also compute the {\em positional} mean-squared error over the positions of the joints. Finally, we report the normalized {\em area under curve (AUC)} for the PCK$_{\rho}$ values, where PCK$_{\rho}$ is a probability for a positional error for a certain landmark to be lower than $\rho$.
For face predictor evaluation we use MPJPE, which is an Euclidean distance between the ground-truth and predicted 3D keypoints. For computing MPJPE we use $68$ SMPL-X head 3D keypoints. For comparing facial expression vectors we employ Mean Squared Error (MSE).

We run the experiments on a single server with AMD Ryzen Threadripper 1900X 8-Core CPU clocked at 3.8GHz and a NVIDIA GeForce RTX 2080Ti GPU.

{\bf Comparison with RGB-based methods.} We start with comparing our approach to the recent RGB-only methods, in order to highlight the benefits of using the depth information. 
We use the recent single frame baselines SMPLify-X~\cite{pavlakos2019expressive}, SPIN~\cite{kolotouros2019learning}, ExPose~\cite{ExPose:2020}. In a sequence-based setup, we compare against the state-of-the-art RGB video-based VIBE~\cite{kocabas2019vibe}. Although our method assumes known shape, other methods estimate the shape alongside with the pose. 

The results on AzurePose, test in~\tab{azure_test_rgb} indicate that our MLP-based model significantly and consistently outperforms the other methods in a single-frame experiment with respect to all metrics by 10-45\%. As our method requires the body shape estimate $\beta$, we use a shape estimated by SPIN~\cite{kolotouros2019learning} as an input to our method. We note that body shape estimation may also benefit substantially from the depth information, but we leave this for future work and use the RGB-based body shape estimate.

In a multiframe experiment, we obtain the results which are better by 5-30\% than the ones of the strongest RGB-based competitor (VIBE~\cite{kocabas2019vibe}) with respect to all the metrics except the local Euler angles on simple sequences, which can be explained by a bias toward complex body poses in the training data. We use $\beta$ estimated by VIBE. 

\begin{figure*}
    \centering
    \includegraphics[width=\textwidth]{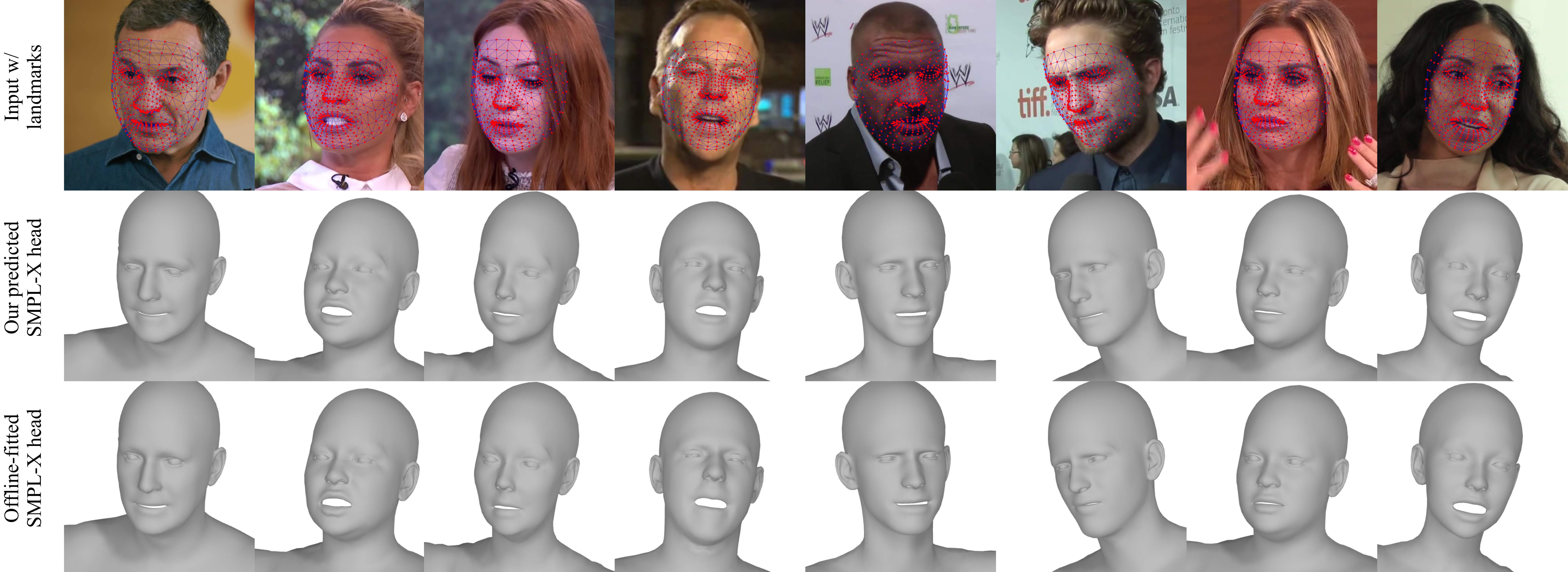}
    \caption{Random samples from VoxCeleb2 test dataset~\cite{chung2018voxceleb2} with face model predictions. Top-to-bottom: input image with overlayed face landmarks~\cite{kartynnik2019real}, predicted SMPL-X model and offline-fitted SMPL-X model (used as ground truth during training, see section~\ref{sect:annotation}).}
    \label{fig:face_examples}
    \vspace{-0.5cm}
\end{figure*}

\begin{table}[]
    \centering
    \resizebox{0.44\textwidth}{!}{
    \begin{tabularx}{0.5\textwidth}{||c|X|X|X|X|X|X||}
         \hline 
         & \multicolumn{2}{c}{Train} & \multicolumn{4}{|c||}{AzurePose test} \\
         \hline
         
         SPL & AZ & AM-K & AUC$\uparrow$ & Euler$\downarrow$  & joint-ang$\downarrow$ & Pos$\downarrow$ \\
         
         \hline
         \hline
         - & + & - & 0.845 & 2.22 & 0.237 & 0.047 \\
         \hline
         + & + & - & 0.853 & 2.286 & 0.222 &  0.044 \\
         
         \hline
         - & + &+ & 0.864 & 2.18 & {\bf 0.207} & 0.04 \\
         \hline
         + & + & + & {\bf 0.867} & {\bf 2.09} & 0.211 & {\bf 0.039} \\
         
         \hline
    \end{tabularx}
        
    }
    \caption{Comparison of the RNN body pose prediction models with or without the SPL layer, trained either only on the AzurePose-train (AZ), or jointly on AzurePose-train and AMASS-K (AM-K). Joint training on two datasets increases pose prediction accuracy.}
    \label{tab:adding_amass_ablation}
    \vspace{-0.5cm}
\end{table}

Overall, we observe a consistent and significant increase in the pose estimation accuracy when depth is used in addition to RGB, see also examples in \fig{rgb_samples}. We conclude from this experiment, that  RGB-D-based human pose estimation still highly relevant for the tasks requiring high accuracy, robustness, and speed. 

\textbf{Comparison with optimization-based approaches.} Next, we compare our model with the single-view RGB-D methods. One of our baselines is the optimization-based offline method 'SMPLify-RGBD', which is a single view modification of the offline model fitting, see \sect{annotation}. Another one is 'SMPLify-RGBD-Online', which is a simplification of the former method to allow for the real-time performance, which does not use VPoser, but relies on the covariance-weighted $l^2$ regularization in the domain of the joint angles as a pose prior, with the covariance matrix computed from the AMASS dataset~\cite{mahmood2019amass}, and uses the $l^2$-norm of the SMPL-X joint motion as a discontinuity cost. All the methods in this comparison received the ground truth body shape $\beta$.

The GRU network in our method serves as a fast replacement for the iterative optimization process, achieving similar accuracy, see \tab{azure_test_rgbd}. With our setting, the feedforward system is 2.5$\times$ faster than 'SMPLify-RGBD-Online' and 25 times faster than 'SMPLify-RGBD'. Comparison to \tab{azure_test_rgb} indicates the increase in accuracy from a better estimate of $\beta$.

On our desktop, the Azure Kinect camera runs at 30 FPS, MediaPipe~\cite{kartynnik2019real} face and our face extractor take 19ms, MinimalHand~\cite{zhou2020monocular} takes 24 ms, our body pose extractor takes 20ms, and composer filter takes 10 ms. The whole system runs at 30 FPS on a desktop computer with a single GPU. To compare, SMPLify-X~\cite{pavlakos2019expressive}, ExPose-X~\cite{ExPose:2020}, SPIN~\cite{kolotouros2019learning} run at less that $1$ FPS, 
VIBE~\cite{kocabas2019vibe} can run at almost $30$ FPS on 2080ti GPU.

\textbf{Ablation study.} In \tab{different_input_data}, we evaluate ablations of our model.
For an RNN-based model we consider four types of input: 3D body landmarks (KP), a shape vector $\beta$, the initial pose $\theta_i^0$ (Init), and the minimal bone-aligning rotations $\mathtt{R}_{i,k}$ (Twists); we compare models predicting rotation increments (Res+), or full rotations (Res-). The AMASS-K dataset has a diverse variety of body shapes, and the best accuracy on this dataset is achieved by a model with highest body shape generalization ability. A model 5 taking $\theta_i^0$ and minimal bone-aligning rotations $\mathtt{R}_{i,k}$ and producing rotation increments achieves lowest errors on AMASS-K, and has high accuracy on AzurePose Test. In \tab{adding_amass_ablation} we show, that  adding AMASS-K with its diverse poses and shapes to the training set helps pose estimation.

\textbf{Face fitting evaluation.} Next, we move on to the evaluation of the jaw pose and facial expression prediction on VoxCeleb2 test dataset~\cite{chung2018voxceleb2}. We compare 3 modifications of our face predictor described in section~\ref{sect:face_and_hands} and report 3 metrics: MPJPE of SMPL-X 3D keypoints, MPJPE of SMPL-X mouth 3D keypoints, and MSE of facial expression vectors. Table~\ref{tab:face_evaluation} summarizes the evaluation results. The first row represents evaluation results for the model, which inputs $2D$ landmarks ($X,Y$-coordinates). Next, we add $Z$-coordinate to the input, which significantly improves all the metrics. Our best model (last row) is additionally trained with $3\times$ weight on SMPL-X 3D mouth keypoints in the loss. We find it important to focus the model on the mouth because jaw pose mainly affects the mouth region.

\begin{table}
    \centering
    \resizebox{0.48\textwidth}{!}{
        \begin{tabular}{||l|c|c|c||}
            \hline
            Model & \begin{tabular}{@{}c@{}}$\downarrow$MPJPE \\ mm\end{tabular} & \begin{tabular}{@{}c@{}}$\downarrow$MPJPE (mouth) \\ mm\end{tabular} & \begin{tabular}{@{}c@{}}$\downarrow$Expression \\ MSE\end{tabular} \\
            \hline
            \hline
            $2D$ landmarks & 2.715 & 3.623 & 2.264 \\
            \hline
            $2.5D$ landmarks & 2.462 & 3.487 & 1.705 \\
            \hline
            $2.5D$ landmarks + mouth loss & \textbf{2.326} & \textbf{3.395} & \textbf{1.478} \\
            \hline
        \end{tabular}
    }
    
    \caption{Face predictor evaluation metrics on VoxCeleb2 test dataset~\cite{chung2018voxceleb2} for 3 modifications of our network. Usage of $Z$-coordinate of input $2.5D$ landmarks  and additional weighing of SMPL-X 3D mouth keypoints in the loss improves metrics.}
    \label{tab:face_evaluation}
    \vspace{-0.6cm}
\end{table}

 \vspace{-0.3cm}
\section{Discussion}
\vspace{-0.2cm}
We have presented the details of the system that estimates the extended pose (including face and hands articulations) from RGBD images or videos in real time. Our systems builds on top of the previously developed components (Kinect pose tracker~\cite{BodyTracker}, MediaPipe face tracker~\cite{kartynnik2019real}, MinimalHand tracker~\cite{zhou2020monocular}) and outputs the result in the popular SMPL-X format~\cite{pavlakos2019expressive}. Importantly, we show that depth-based pose estimation still leads to considerable improvement in accuracy (and speed) compared to RGB-only state-of-the-art approaches. This is, of course, hardly surprising. What is perhaps more surprising given increasing availability of excellent RGB-D sensors is the relatively small interest towards depth-based pose estimation and the lack of available frameworks for extended pose estimation from RGB-D video streams. Our system will, hopefully, address this gap and will be useful to multiple applications in such domains as telepresence and human-computer interaction, where both the simplicity of setup and the accuracy of results are important.

{\small
\bibliographystyle{ieee}
\bibliography{refs}
}
\end{document}